# FarSSiBERT: A Novel Transformer-based Model for Semantic Similarity Measurement of Persian Social Networks Informal Texts

Seyed Mojtaba Sadjadi[1], Zeinab Rajabi[2], Leila Rabiei[3], Mohammad-Shahram Moin[3]

[1] Department of Computer Engineering, Shahrood University of Technology, Shahrood, Iran, mojtabasadjadi@gmail.com

[2] Faculty member in Department of Computer Engineering, Hazrat-e Masoumeh University, Qom, Iran, z.rajabi@hmu.ac.ir

[3] Iran Telecommunication Research Center (ITRC), Tehran, Iran, {l.rabiei, moin}@itrc.ac.ir

*Abstract*

Social networks generate an enormous amount of data through user interactions, which needs to be analyzed and reviewed. One fundamental task for natural language processing is to determine the similarity between two texts and evaluate the extent of their likeness. The previous methods for the Persian language have low accuracy and are unable to comprehend the structure and meaning of texts effectively. Additionally, these methods primarily focus on formal texts, but in real-world applications of text processing, there is a need for robust methods that can handle colloquial and informal texts. This requires algorithms that consider the structure and significance of words based on context, rather than just the frequency of words. The lack of a proper dataset for this task in the Persian language makes it important to develop such algorithms and construct a dataset for Persian text. This paper introduces a new transformer-based model to measure semantic similarity between Persian informal short texts from social networks. In addition, a Persian dataset named FarSSiM has been constructed for this purpose, using real data from social networks and manually annotated and verified by a linguistic expert team. The proposed model involves training a large language model using the BERT architecture from scratch. This model, called FarSSiBERT, is pre-trained on approximately 104 million Persian informal short texts from social networks, making it one of a kind in the Persian language. Moreover, a novel specialized informal language tokenizer is provided that not only performs tokenization on formal texts well but also accurately identifies tokens that other Persian tokenizers are unable to recognize. It has been demonstrated that our proposed model outperforms ParsBERT, laBSE, and multilingual BERT in the Pearson and Spearman's coefficient criteria specific to the similarity measurement tasks. Additionally, the pre-trained large language model has great potential for use in other NLP tasks on colloquial text and as a tokenizer for less-known informal words.

*Keywords* Semantic Similarity Measurement, Colloquial Persian Dataset, BERT, Informal Short Text, Social Networks

## 1. INTRODUCTION

Social networks have become a prevalent part of modern-day social life, with a vast number of users and interactions between them. This creates a vast pool of textual data that can be harnessed for natural language processing tasks such as text summarization [1][2], spam detection [3], news clustering [4], information retrieval [5][6], recommender systems [7], sentiment analysis [8], [9], healthcare [10][11] and more. However, processing these often-short texts presents challenges due to their conversational and informal nature.

GloVe [12] and Word2Vec [13] are neural networks-based word embedding methods that generate static representation vectors by analyzing semantic, syntactic relationships between words. While these methods do not consider the context of the words, transformer-based methods such as the BERT [14] use the context of each word to create a word vector in their equations. This feature help better understanding the meanings of sentences, because like the real language where each word has different meanings, each word would have different embedded vectors.

With the BERT language model and its derived methods, various natural language processing tasks in English and other languages have been accomplished with very good results [15]. Data from various sources including Wikipedia, news, and books have been used to train these large language models in various medical and scientific applications. In the Persian language, ParsBERT [16] is the only model that has been created for the Persian language with official written text from Wikipedia and books.

Social networks texts differ from formal written texts like Wikipedia and news articles in terms of their characteristics. They are brief and use informal grammar, irregular vocabulary, abbreviations, and hashtags. Applying pre-trained linguistic models designed for large-scale regular text sets with formal grammar and regular vocabularies to analyze social networks short texts can be challenging. This is mainly because, the informal nature of social networks texts may not be compatible with these models. To the best of our knowledge, there are no pre-trained language model on a large-scale Persian social network corpus.

In order to address this issue, we train the first Persian language model for social network short texts through training with a 48 GB collection of 104 million Persian tweets which is based on the BERT-base model configuration. We evaluate our model and compare it to strong competitors, namely ParsBERT and multilingual BERT, and laBSE in the downstream task of semantic similarity measurement of Persian informal short texts. Due to the lack of a suitably labeled dataset for this task in the Persian language, we collected a dataset containing 1123 text-pair samples and labeled them with 4 annotations, which we use in model testing and comparing methods. Experiments show that our model performs better than other models. Our contributions are as follows:

- Proposing the first large-scale pre-trained language model for Persian informal short texts (FarSSiBERT).
- Pre-training the first large language model with more than 2 billion tokens in colloquial language from scratch.
- Constructing the first annotated semantic similarity dataset for informal short texts with

real social network data in a standard manner (FarSSiM).

- Training the first dedicated tokenizer for short Persian informal texts, correctly identifying informal and less common words of social networks as tokens.
- Conducting a comprehensive evaluation for analyzing and comparing the proposed method with ParsBERT, laBSE, and multilingual BERT pre-trained models on the downstream task of semantic similarity measurement.

We publicly release our model under the name FarSSiBERT which can be used in Python. We hope that FarSSiBERT can serve as a strong baseline for future research and applications of Persian short texts analytic tasks. It has a high potential for fine-tuning different tasks for colloquial texts.

The rest of the paper is organized as follows. In Section 2, we review current studies related to both language modeling and semantic similarity tasks. The proposed model is introduced in Section 3. Our dataset and its construction approach are presented in Section 4. Evaluations, experimental results, detailed analysis, and discussion are presented in Sections 5 and 6. Ultimately, our work is concluded in Section 7.

## 2. RELATED WORKS

Since this research discusses the task of measuring semantic similarity of informal texts using a language model, we present related works in two parts: Language modeling and Semantic Similarity Task.

### 2.1. Language Modeling

The BERT model was the pioneering work to show that large language models can be efficiently fine-tuned for other natural language processing tasks. This model includes two phases: pre-training and fine-tuning, which in the former phase are trained with Masked Language Modelling (MLM) and Next Sentence Prediction (NSP) without supervision. MLM refers to the prediction of masked words in a sentence, while NSP involves a task for predicting whether the second sentence in a pair of sequences is a true substitute for the first sentence. In the fine-tuning phase, the pre-trained model is tuned for a specific task with labeled data. In the paper, the author tests two types of BERT models with 12 layers (BERT-base) and 24 layers (BERT-large) Transformers [17].

In [18], RoBERTa is presented, which has used 160 GB of textual data to train a large language model. This model is pre-trained with larger batch size and more training steps on longer sequences (512 vs. 128) as compared to BERT. Additionally, the research findings indicate that utilizing the NSP procedure is not useful for end-task performances.

Multilingual BERT [19] is a Transformer-based model that is trained using Wikipedia data for 104 world languages. This model only uses the MLM procedure for pre-training. Although multilingual language models have their advantages, it is evident that monolingual models, specifically trained for the tasks of that language, outperform multilingual ones [20], and the Persian language is no exception [16].

ParsBERT provides a monolingual BERT for the Persian language using data from Wikipedia and Persian books. It fine-tunes its pre-trained model in several tasks, including sentiment analysis, text classification, and named entity recognition, and outperforms multilingual models.

According to research [21], using a pre-trained multilingual model named "laBSE" can significantly decrease the amount of parallel training data needed to achieve good performance by 80%. By combining the best of methods, a model was created that achieves a retrieval accuracy of 83.7% across 109 languages in Tatoeba, exceeding previous methods. This model also performs well on monolingual transfer learning benchmarks. The study published the best embedding model for multilingual sentences in over 109 languages.

In summary, all previous works in the Persian language were based on formal and literary texts. To our knowledge, no model has been trained on the colloquial language of social networks.

*2.2. Semantic Similarity Task*

There is a limited amount of research available on measuring semantic similarity in Persian language due to the absence of appropriate datasets. Majority of the studies conducted have relied on traditional techniques and have not utilized language models. Furthermore, the datasets used are all formal texts that have been sourced from instances of plagiarism or subtitles of films and TV series. Below are some of these studies that have been reviewed.

In [22], Moghadam Emami et al presented a method for semantic similarity measurement on short Persian texts using a deep neural network. Their proposed method consists of three main parts: in the first step, they collect data and form the desired dataset. The next step is pre-processing and text normalization. In the final stage, semantic similarity is done with the help of creating a three-layer neural network model for vocabulary training. For this study, the researchers utilized a dataset containing 35,266 pairs of sentences that were taken from movie and show subtitles. In order to accurately measure semantic similarity between text pairs, they made use of the paraphrased sentences that were available in the movie subtitle database.

Shahabi et al [23] conducted research where they introduced a fuzzy approach to segment Persian text. Their method involves using a semantic similarity measurement between sentences, achieved through fuzzy similarity and fuzzy approximation relation. They first obtain the roots of Persian words and verbs, followed by creating a fuzzy similarity relationship with synonyms. Through this, sentences with similar meanings are calculated using fuzzy proximity relationship. To calculate the proximity of sentences in a fuzzy manner, they use a semantic external knowledge base. Their method serves as a text summarizer that summarizes valuable information from Persian text. However, its accuracy and reliability are limited due to its dependence on an external knowledge base to find synonyms.

In [24], a method for assessing the similarity between categories of two Persian texts is presented. To achieve this, a dataset was compiled, comprising of advertisements posted on an online marketplace. A significant number of text pairs were generated from this dataset, and each was assigned a score between 0 and 3, representing the degree of similarity between them. The study utilized word2vec word embedding vectors to represent words, and deep neural network models to represent text. The neural network was trained to produce a probability distribution vector for the scores of the text pairs. The model achieved an F1 score of 0.9865 using supervised learning.

However, this method has limitations in neural network-based approaches, which can be challenging to interpret and require re-training and re-convergence when reused with new data. Additionally, the dataset used in this study differs from the proposed method of the current research and does not account for the challenges posed by short texts in social networks.

## 3. PROPOSED MODEL

This section outlines the methodology used to create the proposed model. The model is created by pre-training a large language model that follows the BERT-base architecture. The pre-trained model is then utilized to extract word embedding vectors in the downstream task of measuring semantic similarity in Persian informal short texts.

In Figure 1, the training steps of the proposed model are illustrated. The process begins with extracting the social network data texts followed by necessary pre-processing. The cleaned data is then broken into appropriate tokens using the transformer architecture and fed to the model for training. The transformer model consists of 12 layers, and its weights are utilized to extract the embeddings of both sentences and words.

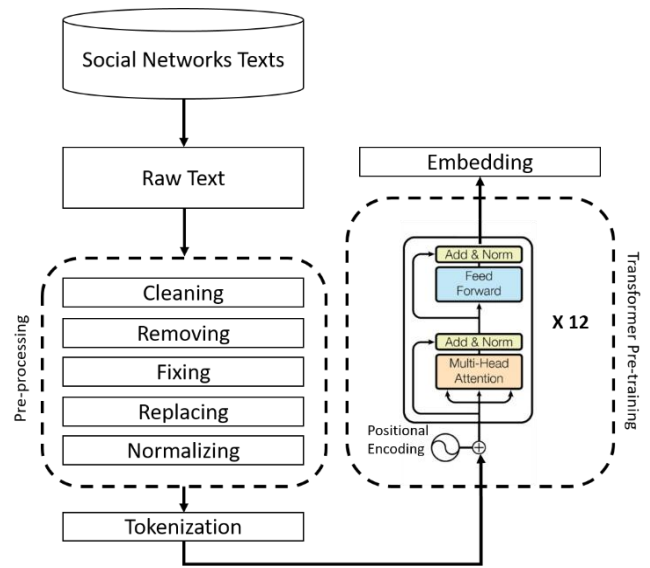

**Figure 1)** Proposed model training

### 3.1. Training data

Earlier, pre-trained models like ParsBERT were trained using formal data from sources like Wikipedia and books. Although these models have their strengths, they have limitations in detecting informal language tokens and the irregular structure of short texts. To overcome this, we trained our model using data from social networks that feature colloquial and informal language. We have gathered a vast collection of Persian-language tweets from Twitter, consisting of over 104M tweets from 250 million between the years 2019 to 2023. Our training dataset size is about **38GB**, with **104551461** unique tweets and **2.18** billion tokens. It includes a diverse range of tweets and threads on various topics from Persian speakers worldwide, not just limited to a specific region.

*3.2. Pre-processing*

Once the training dataset is gathered, it is crucial to perform specific pre-processing steps, such as cleaning, removing, replacement, and normalization, to transform the dataset into a compatible format. Table 1 lists these essential steps. In semantic methods processing, replacement is a crucial step as it removes tokens that do not contribute to the meaning of the text. For example, without replacing the @username with the term USERNAME, the word of the person's name mentioned in the middle of tweet, may add other meanings to the tweet, which will cause the model to misunderstand. In addition, because Emojis carry meaning and to preserve their meaning, a literal equivalent has been substituted for each emoji in the text.

**Table 1)** Persian tweets pre-processing steps

| Method | Items |
|---|---|
| **Clean** | HTML tags, JavaScript, CSS |
| **Remove** | Double spaces, Line breaks, Email, Phone numbers, Unwanted ASCII, Weird Unicode, # |
| **Fix** | Unicode |
| **Replace** | @username => USERNAME<br>url => URL<br>Arabic characters => Persian characters<br>Emoji => Meaning word |
| **Normalize** | Using HAZM[1] |

*3.3. Pre-training*

The proposed model is based on a transformer and trained according to BERT-base architecture, which includes 12 hidden layers, 12 attention heads, and 768 hidden sizes. The pre-training of the BERT model consists of two processes according to the type of data one or both processes can be used in model training. In this study, the Masked Language Model (MLM) is used for pre-training for the following reasons:

- Informal data do not follow a specific structure for sentences and it is impossible to separate sentences.
- Short texts are generally single sentences
- There is no suitable Persian-labeled data for the Next Sentence Prediction (NSP) procedure of informal texts.

Likewise, a masked language model is used to train the model to predict masked tokens using cross-entropy loss. Out of the total number of tokens, 15% of them are masked by the [MASK] token, and the model is trained by trying to predict these masked words. The size of the hidden layer is 768, the batch size is 32, the length of the input strings is 512, and the learning rate is 1e-12. We optimize the model using Adam [25] and pre-train FarSSiBERT for 5 epochs in about 23 days in about 16M steps across one A100 GPU (40GB RAM). Figure 2 shows the loss curve for model training in 5 epochs.

---

[1] https://www.roshan-ai.ir/hazm/

In this model, a sub-word-based tokenizer

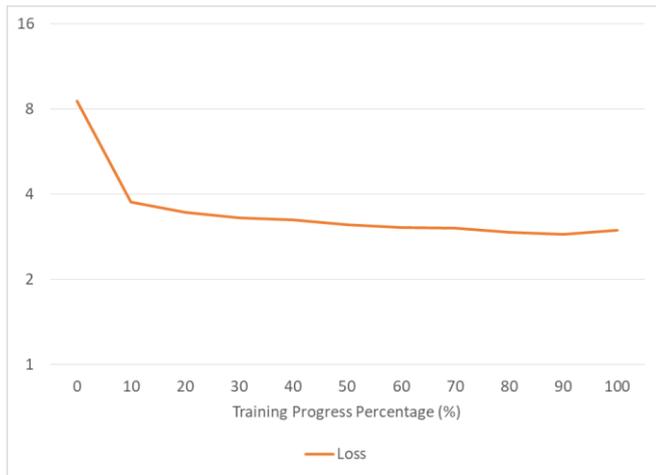

**Figure 2)** The loss curve of the pre-trained model in the training process

(wordpiece algorithm) is used, which includes specific tokens of the BERT model named [PAD], [UNK], [CLS], [MASK] [SEP], and suffixes are marked with [##] separates from the original word as a token. This tokenizer is trained by our training dataset and contains 100K vocab list, which is the first tokenizer trained for informal Persian short texts.

*3.4. Semantic Similarity Measurement*

As previously mentioned, we utilize the pre-trained model to measure the semantic similarity of short informal Persian texts. The process is illustrated in Figure 3, which outlines the use of FarSSiBERT. Two short texts are inputted into the pre-trained model after necessary pre-processing and tokenization. Token embeddings are extracted using the average of the last 4 layers of FarSSiBERT, and sentence embeddings are created through a pooling operation. We experimented with three pooling strategies: the CLS token output, computing the mean of all output vectors (Mean strategy), and computing the maximum time of the output vectors (Max strategy). The Mean strategy is the default configuration.

After extracting the embedding vectors of the sentences, the degree of similarity between the two vectors is calculated by the cosine distance and a similarity score between [0 , 5] is given to these two texts. Due to left-to-right and right-to-left text processing, the proposed model has well understood and trained the semantic similarities between words in the context, which is suitable for semantic tasks.

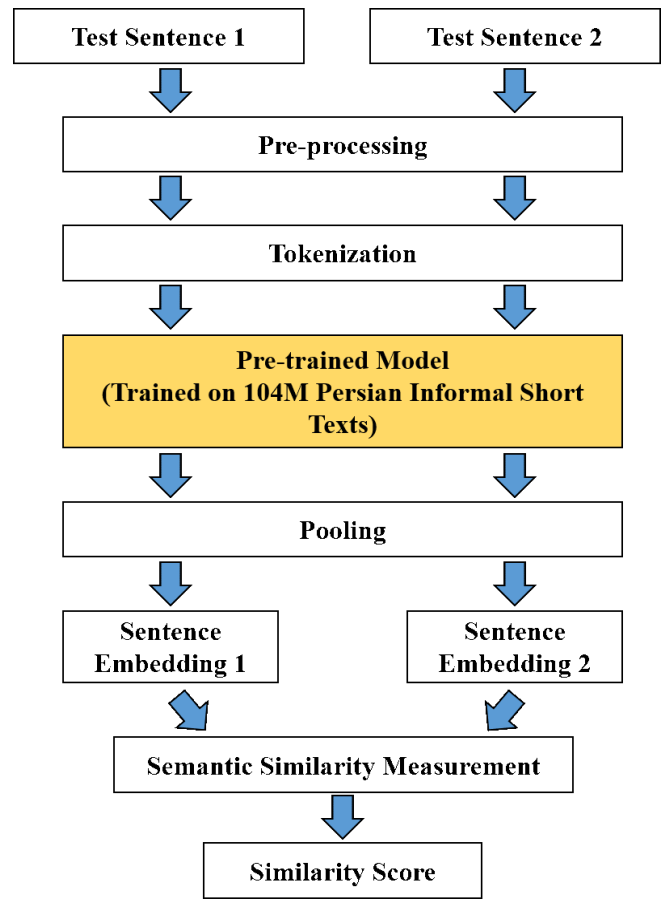

**Figure 3)** Semantic Similarity Measurement using FarSSiBERT

*3.5. Metrics*

Pearson's and Spearman's coefficients are common measures to measure the semantic similarity of texts [26], [27], which are described below.

### 3.5.1. Pearson coefficient

The Pearson coefficient is a measure of how closely the results of a given test match up with human judgments. It is computed as Eq.1.

$$r = \frac{\sum(x_i - \bar{x})(y_i - \bar{y})}{\sqrt{(x_i - \bar{x})^2 - (y_i - \bar{y})^2}} \quad (1)$$

Where $x_i$ = values of the x-variable in a sample,

$\bar{x}$ = mean of the values of the x-variable,

$y_i$ = values of the y-variable in a sample, $\bar{y}$ = mean of the values of the y-variable.

### 3.5.2. Spearman's Coefficient

The classification is determined by measuring the similarity between sentences and comparing it to the similarity determined by human judgments. Spearman is calculated as follow:

$$p = 1 - \frac{6 \sum d_i^2}{n(n^2 - 1)} \quad (2)$$

Where $di$ corresponds the difference of the ranks of $x_i$ and $y_i$, and $n$ corresponds to the pair sentence number.

### 3.5.3. Mean Square Error

Also, for error analysis, the Mean Square Error (MSE) criterion is used, the equation of which is given below:

$$MSE = \frac{1}{n}\sum_{i=1}^{n}(l_{desired} - l_{predicted})^2$$

Where $l_{desired}$ is real label and $l_{predicted}$ is the label predicted by the model.

## 4. DATASET CONSTRUCTION

### 4.1. Persian dataset for semantic similarity

Mashhadi Rajab and Shamsfard [28][29] created a dataset for evaluating plagiarism detection systems using scientific texts. The collection contains 11089 documents that are prepared from articles in the fields of computer science and electrical engineering. There are 11603 cases of plagiarism in this collection. Only one percent of this corpus can be used to measure semantic similarity. Also, they are official scientific documents and due to the focus on the field of computer science and electricity, this corpus includes a smaller range in terms of data variety.

Minaei and Niknam [29][30] collected a set of 3218 documents and 2308 cases of plagiarism to find similar parts of two given documents. They used Wikipedia articles to build their collection and a set of text operations to generate instances of plagiarism such as copying text from a source document or replacing words with their synonyms.

Sharif Abadi [31] prepared a collection to detect plagiarism from Persian scientific texts containing cases of plagiarism. 4707 articles used in this collection have been collected from the fields of humanities, sciences, veterinary medicine, agriculture and natural sciences, engineering, art, and architecture. The collection contains 5000 plagiarism cases of various lengths. Semantic plagiarism cases are only 20% of this collection, equivalent to 1172 cases.

The Abnar dataset [32] includes a collection of novels translated into Persian. There are 2470 documents and 12061 cases of plagiarism in this collection, which has the highest number of plagiarism cases among other collections. Only 15% of this collection is dedicated to semantic plagiarism.

Khoshnava Taher [33] was collected using 1611 pairs of documents containing random plagiarism, 208 document pairs containing unambiguous plagiarism, 147 pairs with simulated plagiarism, and 783 pairs without plagiarism. The total number of documents is 3088 including 1525 suspicious documents and 1563 reference documents. Five different obfuscation strategies are used to gather this corpus; however, this set is not suitable for neural network training because the number of plagiarism cases is limited. In addition, no details are provided on the number of existing semantic plagiarism cases.

Perpada [34] is a Persian paraphrase dataset collected from users' input into a plagiarism detection system. A large collection of original and rewritten sentences compiled from the Hamtajo website system which is a Persian plagiarism detection system where users try to analyze text reuse in their documents by rewriting and resubmitting manuscripts. The final paraphrasing dataset of Perpada contains 2446 pairs of sentences with a length between 50-100 letters.

To our understanding, Persian studies on semantic similarity have only been conducted on datasets related to plagiarism, which aim to identify text matches. These datasets primarily consist of lengthy academic and formal documents. However, no research has been carried out on social network data in the Persian language that comprises informal short texts. In order to assess and validate our proposed model, we require a labeled dataset. As there is a lack of such datasets, we have taken the initiative to create a public dataset of short Persian texts on Twitter named ***FarSSiM*** (Farsi Semantic Similarity Measurement). In the following sections, we will elaborate on the methods we employed to collect and label this dataset.

*4.2. FarSSiM*

To measure semantic similarity, datasets consist of pairs of paraphrased text which are labeled to indicate the level of similarity between them. Finding these kinds of samples with equal meaning requires identifying paraphrases. A method to do this is identifying common words between tweets on a particular topic and time. Once the pairs of paraphrased tweets are identified, expert annotators compare them semantically and assign a score as a label. The final score for each pair is determined by the average of the annotators' votes and inserted into the dataset [35][36][37].

*4.2.1. Identification of candidate pairs*

The required data is collected by a web crawler from Persian Twitter at specific times when there is a popular topic. To identify the candidate pairs, all the data was pre-processed, and items that did not contribute to the semantic similarity measurement of the texts (tweets with a length of less than 4 words) were removed (values greater than and less than 4 were tested and the best result was obtained with the value of 4) [38][39]. In the next step, all possible pairs of tweets were considered candidate pairs, then filtered according to the following criteria. A candidate pair is eliminated if only one of the following conditions is met:

1. Do not be in a trend.
2. The time interval between inserting two tweets is more than one day.
3. Two tweets overlap in less than 5 words.

*4.2.2. Semantic Similarity Scoring*

Rather than categorizing text pairs as either paraphrasing or non-paraphrasing, we assign a more accurate semantic similarity score. Our team followed the guidelines outlined for the semantic similarity task, and annotators scored each text pair directly based on the criteria outlined in Table 2.

**Table 2)** Semantic Similarity scores for candidate paraphrase pairs

| Score | Meaning |
|---|---|
| **5- Identical** | Completely equivalent; mean the same |
| **4- Close** | Mostly equivalent; some small details differ |
| **3- Related** | Roughly equivalent; some important information differs/missing |
| **2- Context** | Not equivalent, but share some details |
| **1- Somewhat related** | Not equivalent, but are on the same topic |
| **0- Unrelated** | On different topics |

We hired four Persian-speaking experts as annotators and provided them with a set of guidelines to score the data. Our instructions included examples and explanations of similarity scores to ensure consistency. The label for each dataset entry was determined by averaging the scores of the four annotators. All Persian annotators are native speakers who possess proper knowledge and understanding of social media, Persian grammar, and background knowledge of natural language processing. Furthermore, all annotated documents were reviewed by an experienced annotator.

In constructing the dataset, the paraphrases in the texts of the social network posts were selected based on the following criteria:

1) Posts from various domains and topics were collected without limiting to trends in one topic. 2) Advertising posts were excluded from the collection. 3) To maintain diversity, posts from both influential users with a high number of likes and ordinary users with low influence were selected. We also included the standard deviation and variance of their scores in the dataset for reference. Three examples of the instances available in the FarSSiM are given in Table 3. For more information, we have provided the English translation of the dataset text alongside the Persian examples.

**Table 3)** Three examples of FarSSiM

| Tweet1 | Tweet2 | 1st | 2th | 3th | 4th | *Avg* | *SD* | *Var* |
|---|---|---|---|---|---|---|---|---|
| نایب‌رئیس فدراسیون فوتبال بازداشت شد نایب‌رئیس فدراسیون فوتبال به‌دلیل ارتباط با پرونده کلاهبرداری رمزارز جعلی کینگ‌مانی بازداشت شد. (**Meaning:** The vice president of the football federation was arrested. The vice president of the Football Federation was detained due to the connection with the King money fake cryptocurrency fraud case.) | صدا و سیما به نقل از یک منبع آگاه از بازداشت نایب رییس فدراسیون فوتبال از سوی وزارت اطلاعات در خصوص پرونده «کلاهبرداری رمز ارز کینگ مانی» خبر داد. (**Meaning:** IRIB, quoting an informed source, announced the arrest of the vice president of the Football Federation by the Ministry of Information regarding the "King Money cryptocurrency fraud" case.) | 4 | 4 | 5 | 4 | **4.25** | 0.5 | 0.25 |
| ادامه حواشی باشگاه لاکچری بدنسازی؛ مدیر کل ورزش استان تهران: پلمب می‌شود در حالی که حواشی رونمایی از باشگاه بدنسازی «متافیتنس» همچنان ادامه دارد، مدیر کل ورزش استان تهران از پلمب شدن این مجموعه خبر داد. (**Meaning:** Continuation of the margins of the luxury bodybuilding club; General Director of Sports of Tehran Province: It will be sealed While the unveiling of "Metafitness" gymnasium is still going on, the General Director of Sports of Tehran Province announced that this complex will be sealed.) | جنجال بر سر افتتاح باشگاه متا در خیابان فرشته تهران؛ وزارت ورزش می‌گوید تعطیل خواهد شد، منتقدان می‌گویند صاحبان ثروتمند باشگاه مشکل مجوز را حل خواهند کرد. (**Meaning:** Controversy over the opening of Meta Club on Fereshteh Street in Tehran; The sports ministry says it will close, critics say the club's wealthy owners will solve the licensing problem.) | 3 | 3 | 4 | 4 | **3.5** | 0.57 | 0.33 |
| شاخص آلودگی هوای تهران به بالای ۲۰۰ رفته و در شرایط خیلی‌ناسالم قرار گرفته است. (**Meaning:** Tehran's air pollution index has risen above 200 and is in very unhealthy conditions.) | شرکت کنترل کیفیت هوای تهران وضعیت هوای پایتخت ایران را "بسیار ناسالم و بنفش" اعلام کرد. میانگین شاخص کیفیت هوای پایتخت هم‌اکنون ۲۰۱ است و مدارس ابتدایی نیز تعطیل اعلام شده است. (**Meaning:** Tehran Air Quality Control Company announced the air condition of the capital of Iran as "very unhealthy and purple". The average air quality index of the capital is 201 and elementary schools are also closed.) | 3 | 3 | 4 | 4 | **3.5** | 0.57 | 0.33 |

In this table, 1st, 2st, 3st and 4st are the scores of the first to fourth annotators, $SD$ is the standard deviation, $Var$ is the variance, and the $Average$ is the average score. The FarSSiM dataset comprises 1123 samples, and their label distribution is illustrated in Figure 4. This dataset is the first-ever collection in Persian that can serve the purpose of semantic similarity analysis of short informal texts and detecting paraphrases in the Persian language.

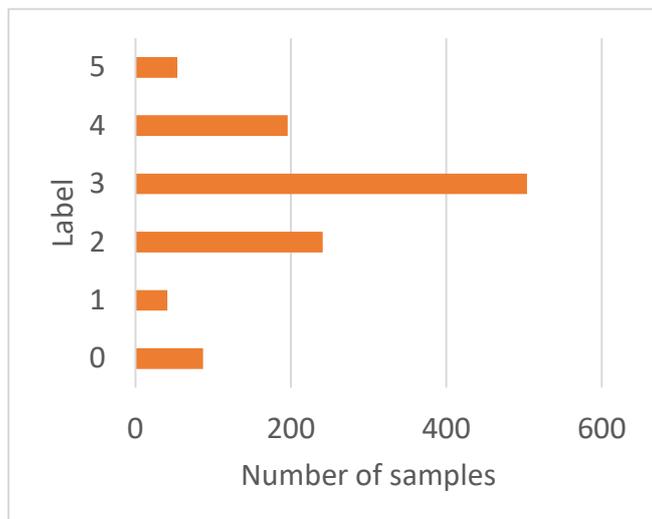

**Figure 4)** Distribution of FarSSiM dataset labels

As seen in Figure 4, label "3" has the most frequency among FarSSiM labels. Because it is the boundary between paraphrasing and non-paraphrasing of two tweets, and the more the model is evaluated in this label, the better the performance of the model is identified.

## 5. EVALUATION

We compare the performance of FarSSiBERT with baselines on the downstream NLP task of informal short-text semantic similarity measurement, using two datasets.

### 5.1. Dataset

As discussed in Section 4.2, there was no adequate Persian dataset available for our research task, so we had to gather and prepare our own test dataset. However, we did come across another dataset called FarSick [40], which consists of 10,000 pairs of short texts in Persian. This dataset was originally in English and was later translated into Farsi. The text pairs in this dataset are written in formal language and lack colloquial and informal data. Each sample is assigned a numerical label ranging from 0 to 5.

To assess the performance of the FarSSiBERT model, we used not only the dataset we collected for this research (FarSSiM) but also the FarSick dataset to evaluate how well the model performs on formal short texts.

### 5.2. Baselines

multilingual BERT model [19]: a pre-trained model that was trained in 104 languages of the world and used the data available in Wikipedia for training.

ParsBERT [16]: As mentioned in previous works, the ParsBERT model is trained with about 48 million Persian data and fine-tuned in various tasks.

laBSE [21]: A large language model based on the BERT architecture, trained for 104 world languages with official data from Wikipedia and other sources.

FarSSiBERT-6M: Since the number of training data has a significant effect on the performance of the model, a smaller model of the proposed method is trained with fewer data (6M) to be compared with the main model of this research, FarSSiBERT-104M.

Due to the limitations and lack of many models for comparison in the Persian language, these models are used to evaluate and compare with the proposed model.

*5.3. Tokenizers Comparison*

In this study, a new tokenizer is introduced that can identify new informal tokens and outperforms other tokenizers. We compare the FarSSiBERT tokenizer with the ParsBERT tokenizer and evaluate their performance on both official and unofficial data types, which differs from previous studies. Examples in Table 4 highlight how the FarSSiBERT tokenizer model compares to the ParsBERT model. The ParsBERT model's tokenizer is incompetent to recognize informal words as independent tokens, leading to these words being broken down into smaller tokens. This can cause the model to misunderstand sentence meanings. In contrast, the FarSSiBERT model identifies each of these words as meaningful tokens, enabling the model to understand them accurately.

**Table 4)** Comparison of some examples of FarSSiBERT model tokenizer with ParsBERT model

| Sentence | ParsBERT tokenizer | FarSSiBERT tokenizer |
|---|---|---|
| بیا فردا برویم کتابخونه | ['بیا', 'فردا', 'برویم', 'کتاب', '##خونه'] | ['بیا', 'فردا', 'برویم', 'کتابخونه'] |
| ظاهرا هنوز برق وصل نشده و ایشون هم ندیده جلو پاشو | ['ظاهرا', 'هنوز', 'برق', 'وصل', 'نشده', 'و', '[UNK]', 'هم', 'ند', '##ید', '##ه', 'جلو', 'پاشو'] | ['ظاهرا', 'هنوز', 'برق', 'وصل', 'نشده', 'و', 'ایشون', 'هم', 'ندیده', 'جلو', 'پاشو'] |
| این رفتارا چیه که نشون میدین، گفتمش ببینه خیابونو | ['این', 'رفتار', '##ا', 'چیه', 'که', 'نشون', 'میدین', '،', 'گفتم', '##ش', 'ببینه', 'خیابون', '##و'] | ['این', 'رفتارا', 'چیه', 'که', 'نشون', 'میدین', '،', 'گفتمش ببینه خیابونو'] |

*5.5. Results*

The proposed model is evaluated in the downstream task of semantic similarity of informal Persian short texts in social networks. Tables 5 and 6 respectively show the values of the scores obtained by the models in the FarSick and FarSSiM datasets. For each pre-trained model, Pearson and Spearman correlation values are calculated.

**Table 5)** Performance scores on the FarSick

| Model | Pearson | Spearman |
|---|---|---|
| **FarSSiBERT-104M** | **0.616** | **0.5829** |
| laBSE [21] | 0.614 | 0.5660 |
| ParsBERT [16] | 0.592 | 0.563 |
| multilingual BERT [19] | 0.556 | 0.534 |
| FarSSiBERT-6M | 0.551 | 0.526 |

It is evident that the FarSSiBERT-104M model outperforms all other models. Despite the fact that the instances in the FarSick dataset are of a formal language type, it was anticipated that the ParsBERT model, which was trained with this type of formal data, would outperform the other models. However, the proposed model performed about 3% better than ParsBERT. Furthermore, the model outperformed the laBSE model, which is trained with formal text. The multilingual BERT and FarSSiBERT-6M models ranked lower and showed the lowest performance, indicating that the number of training data has a significant impact on the model's performance.

In another experiment, the results of comparisons with other methods are shown in Table 6. The degree of similarity of the text pairs in the dataset collected in this research (FarSSiM) has been predicted by different pre-trained models and compared with the actual labels, in order to evaluate the performance of the models.

Table 6) Performance scores on the FarSSiM

| Model | Pearson | Spearman |
|---|---|---|
| FarSSiBERT-104M | **0.770** | **0.643** |
| FarSSiBERT-6M | 0.740 | 0.621 |
| laBSE [21] | 0.725 | 0.643 |
| ParsBERT [16] | 0.704 | 0.624 |
| multilingual BERT [19] | 0.618 | 0.480 |

After analyzing the data in Table 6, it is clear that the models developed in this research outperformed other models. The FarSSiBERT-104M, with a Pearson correlation of 0.770 and a 7% difference from the ParsBERT model, demonstrates the effectiveness of using appropriate training data and a dedicated tokenizer. Furthermore, the ParsBERT model uses the NSP procedure in its training process and has been fine-tuned for various tasks. However, this research's model achieved better results without this procedure, indicating its high potential for a broad range of natural language processing tasks. Additionally, even the FarSSiBERT-6M performed better than laBSE, ParsBERT, and multilingual BERT in Pearson's benchmark and reported superior results.

## 6. DISCUSSION

The previous section compared the model's performance with baseline models. The results are influenced by how the model performed with different data inputs and labels. To gain a better understanding of these findings, an analysis of the model's errors is conducted. The proposed model is evaluated alongside baseline models to predict appropriate labels for similarity measurement. The error between the predicted label and the original label is calculated for each data sample, and the results can be found in Table 7. This table displays the mean square error for each label across different models in the FarSSiM and FarSick datasets. For instance, the column (0<L <1) presents the mean square error for data samples with labels between 0 and 1.

*FarSSiM error analysis:* Based on the error values for different labels in the FarSSiM dataset, it is evident that, in general, all models perform better with larger labels (indicating more similarity of text pairs) than with smaller labels. The models show the lowest error in labels between 4 and 5. For data with labels between 0 and 1, the proposed model has the least error, with an MSE value of 1.23, outperforming other models. On the other hand, the multilingual BERT model displays the worst performance with an error of 5.09, indicating its poor performance in dealing with non-paraphrased data for measuring semantic similarity.

The labSE model has slightly less error in data with labels between 1 and 2, but the proposed model maintains good performance throughout the rest of the labels, outperforming other models. Figure 4 shows that the majority of labels in the FarSSiM dataset are placed in labels between 2 and 4. This distribution and Table 7 values reflect the effectiveness of the proposed model in detecting semantic similarity in borderline conditions between paraphrasing and non-paraphrasing of two texts. Accordingly, according to Table 7, the proposed model has the least error in these labels and can detect semantic similarity between two texts better than other models.

*FarSick error analysis:* The error values in the FarSick dataset decrease as the labels increase, suggesting improved model performance for similar texts. There are no labels between 0 and 1 in this dataset, so the values in this column are null. The proposed model demonstrates lower error rates across all labels and better identifies similarities in informal short texts. Following the proposed model, the labSE model is ranked next, with the ParsBERT and multilingual BERT models following. The table values indicate that error values greater than 1 have been obtained for data labeled between 2 and 4 -representing the boundary conditions of paraphrasing and non-paraphrasing of two texts- indicating higher errors compared to the FarSSiM dataset. This demonstrates the good quality of the FarSSiM dataset, as most models displayed lower errors for these values.

After analyzing errors in various labels and evaluating the performance of the proposed model, it can be concluded that using appropriate data and a transformer-based model that can analyze texts bidirectionally and identify the semantic relationship between tokens helps measure the semantic similarity of informal short texts from social networks. The use of a special tokenizer, proper pre-processing, data cleaning methods, and accurate substitutions significantly contribute to this result. Unlike the baseline models, the proposed models only utilized the masked language modeling and pre-training procedure for model training. If higher-quality data can be obtained for fine-tuning, they are expected to exhibit improved accuracy and performance.

**Table 7)** MSE values for different labels of datasets

| Dataset | Model | MSE | | | | |
|---|---|---|---|---|---|---|
| | | 0<L<1 | 1<L<2 | 2<L<3 | 3<L<4 | 4<L<5 |
| FarSSiM | FarSSiBERT-104M | **1.23** | 1.02 | **0.85** | 0.60 | **0.38** |
| | FarSSiBERT-6M | 3.25 | 2.50 | 1.59 | **0.58** | 0.39 |
| | laBSE [21] | 1.25 | **0.94** | 0.93 | 0.65 | 0.51 |
| | ParsBERT [16] | 1.88 | 3.06 | 1.37 | 0.64 | 0.42 |
| | multilingual BERT [19] | 5.09 | 4.11 | 1.92 | 0.68 | 0.41 |
| FarSick | FarSSiBERT-104M | - | **0.85** | **1.06** | **1.02** | **0.33** |
| | FarSSiBERT-6M | - | 1.25 | 1.19 | 1.11 | 0.82 |
| | laBSE [21] | - | 0.96 | 1.17 | 1.19 | 0.51 |
| | ParsBERT [16] | - | 1.67 | 1.16 | 1.10 | 0.75 |
| | multilingual BERT [19] | - | 1.95 | 1.12 | 1.05 | 0.84 |

# 7. CONCLUSION

We have presented the first large-scale language model FarSSiBERT, which is pre-trained for short informal Persian texts. We demonstrate the utility of FarSSiBERT by showing outperforming its baseline ParsBERT, laBSE, and multilingual BERT and helping provide better performance than previous SOTA models for the downstream task of semantic similarity. Also, due to the fact that there is no suitable dataset for the task of semantic similarity measurement of informal Persian short texts, we collected and labeled the FarSSiM dataset containing 1123 text pairs for testing the model.

Our pre-trained FarSSiBERT was developed using over 104 million Persian tweets and over 2.18 billion tokens. It trained in five epochs and was evaluated based on Pearson and Spearman criteria for the semantic similarity test. The tokenizer used in this model is highly effective at identifying colloquial and informal tokens. To facilitate the use of the FarSSiBERT in other natural language processing tasks, a Python package has been created and made publicly available, along with the FarSSiM dataset.